\pdfoutput=1
\documentclass{article}
\usepackage{graphicx}
\usepackage{subfigure}
\graphicspath{ {images/} }
\usepackage{multirow}
\usepackage{natbib}
\usepackage{algorithm}
\usepackage{algorithmic}
\usepackage{hyperref}

\usepackage[accepted]{icml2013}
\icmltitlerunning{Image-Text Multi-Modal Representation Learning by Adversarial Backpropagation}

\begin{document}

\twocolumn[
\icmltitle{Image-Text Multi-Modal Representation Learning by Adversarial Backpropagation}

\icmlauthor{Gwangbeen Park}{bobmarley66@kaist.ac.kr}
\icmlauthor{Woobin Im}{iwbn@kaist.ac.kr}
\icmladdress{Korea Advanced Institute of Science and Technology (KAIST), Daejeon, Republic of Korea}
\icmlkeywords{deep learning, adversarial, multi modal, search}

\vskip 0.3in
]

\begin{abstract} 
We present novel method for image-text multi-modal representation learning. In our knowledge, this work is the first approach of applying adversarial learning concept to multi-modal learning and not exploiting image-text pair information to learn multi-modal feature. We only use category information in contrast with most previous methods using image-text pair information for multi-modal embedding.

In this paper, we show that multi-modal feature can be achieved without image-text pair information and our method makes more similar distribution with image and text in multi-modal feature space than other methods which use image-text pair information. And we show our multi-modal feature has universal semantic information, even though it was trained for category prediction. Our model is end-to-end backpropagation, intuitive and easily extended to other multi-modal learning work.
\end{abstract} 

\section{Introduction}
\label{Introduction}
\begin{figure}[ht]
    \includegraphics[width=\linewidth]{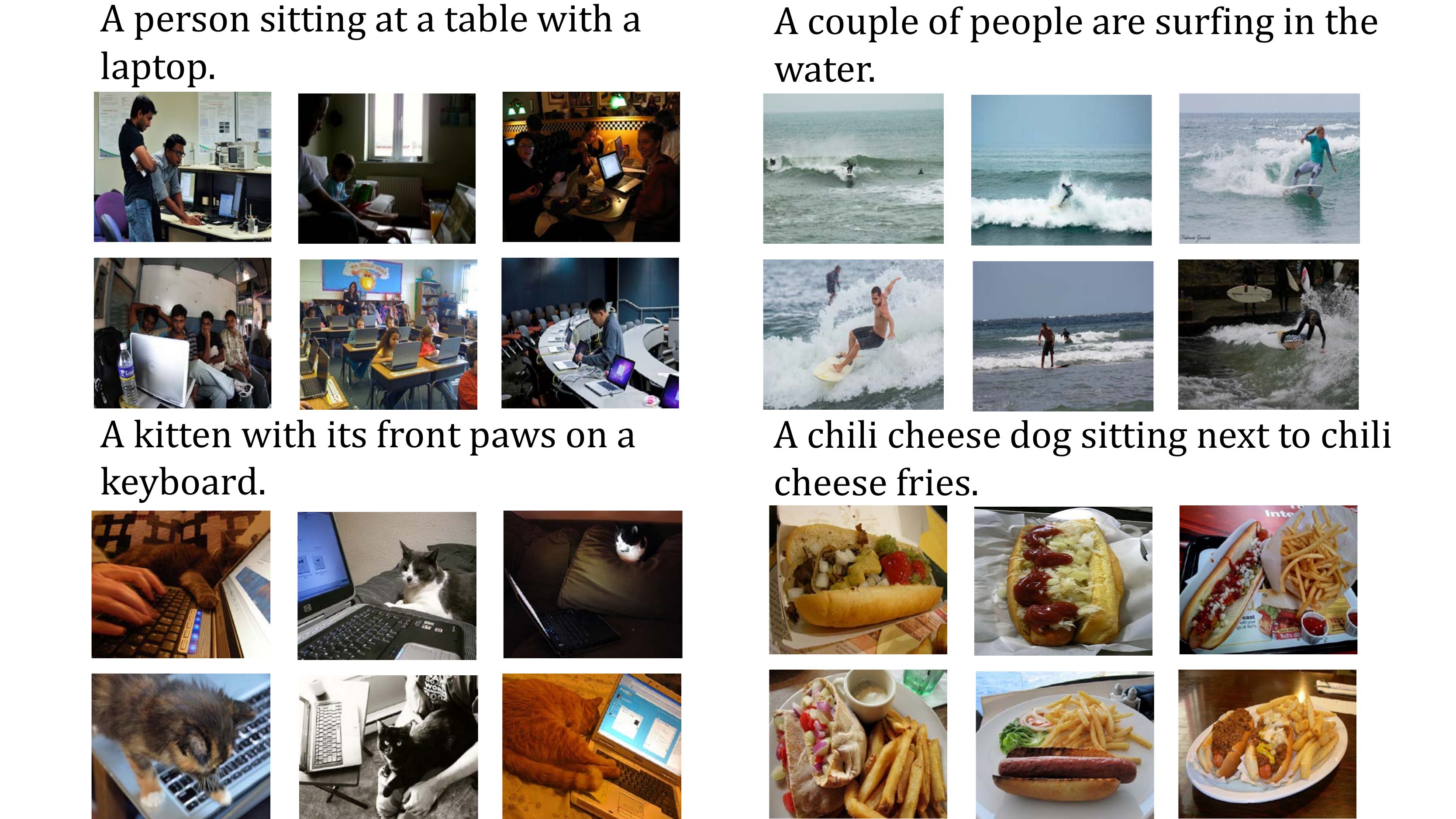}
    \caption{Sentence to Image search example from our search system with MS COCO dataset. (upper sentence is query, below images are search result.}
    \label{fig:intro_search_ex}
\end{figure}
Recently, several deep multi-modal learning tasks have emerged. 
There are image captioning \cite{vinyals2015show}, text conditioned image generation \cite{DBLP:conf/icml/ReedAYLSL16}, object tagging \cite{karpathy2015deep}, text to image search \cite{Wang_2016_CVPR}, and so on. For all these works, how to achieve semantic multi-modal representation is the most crucial part. \newline
Therefore, there were several works for multi-modal representation learning \cite{srivastava2012multimodal,frome2013devise,sohn2014improved,Wang_2016_CVPR}. And all of these works require image-text pair information. Their assumption is, image-text pair has similar meaning, so if we can embed image-text pair to similar points of multi-modal space, we can achieve semantic multi-modal representation. \newline
But pair information is not always available in several situations. Image and text data usually not exist in pair and if they are not paired, manually pairing them is an impossible task. But tag or category information can exist separately for image and text. And also, does not require paired state and can be manually labeled separately. \newline
And learning multi-modal representation from image-text pair information can be a narrow approach. Because, their training objective focuses on adhering image and text in same image-text pair and doesn't care about adhering image and text, that are semantically similar, but in different pair. So some image and text can have not similar multi-modal feature even though they are semantically similar. In addtion, resolving every pair relations can be a bottleneck with large training dataset. \newline
To deal with above problems, for multi-modal representation learning, we bring concept from ganin's work\cite{icml2015_ganin15} which does unsupervised image to image domain adaptation by adversarial backpropagation. They use adversarial learning concept which is inspired by GAN (Generative Adversarial Network)\cite{goodfellow2014generative} to achieve category discriminative and domain invariant feature. We extend this concept to image-text multi-modal representation learning. \newline
We think image and text data are in covariate shift relation. It means, image and text data has same semantic information or labelling function in high level perspective but they have different distribution shape. So we regard, multi-modal representation learning process is adapting image and text distribution to same distribution and retain semantic information at the same time.\newline
In contrast with previous multi-modal representation learning works, we don't exploit image-text pair information and only use category information. Our focus is on achieving category discriminative, domain (image, text) invariant and semantically universal multi-modal representation from image and text. \newline
With above points of view, we did multi-modal embedding with category predictor and domain classifier with gradient reversal layer. We use category predictor for achieving discriminative power of multi-modal feature. And using domain classifier with grdient reversal layer, which makes adversarial relationship with embedding network and domain classifier, for achieving domain (image, text) invariant multi-modal feature. Domain invariant means image and text have same distribution in multi-modal space.\newline  
We show that our multi-modal feature distribution is well mixed about domain, which means image and text multi-modal feature's distributions in multi-modal space are similar, and also well distributed by t-SNE\cite{van2014accelerating} embedding visualization. And comparison classification performance of multi-modal feature and uni-modal (Image only, Text only) feature shows, there exists small information loss within multi-modal embedding process and still multi-modal feature has category discriminative power even though it is domain invariant feature after multi-modal embedding. And our sentence to image search result (Figure \ref{fig:intro_search_ex}) with multi-modal feature shows our multi-modal feature has universal semantic information, which is more than category information. It means, within multi-modal-embedding process, extracted universal information from Word2Vec\cite{mikolov2013distributed} and VGG-VeryDeep-16 \cite{DBLP:journals/corr/SimonyanZ14a} is not removed. \newline
In this paper, we make the following contributions. First, we design novel  image-text multi-modal representation learning method which use adversarial learning concept. Second, in our knowledge, this is the first work that doesn't exploit image-text pair information for multi-modal representation learning. Third, we verify image-text multi-modal feature's quality in various perspectives and various methods.\newline
Our approach is much generic as it can be easily used for any different domain (e.g. sound-image, video-text) multi-modal representation learning works with backpropagation only. 

\section{Related Works} 
\subsection{Multi-Modal Representation}
Several works about image-text multi-modal representation learning have been proposed over the recent years. Specific tasks are little bit different for each work, but these works' crucial common part is achieving semantic image-text multi-modal representation from image and text.  \newline
Image feature extraction and text feature extraction method are different with each work. But almost they commonly use image-text pair information to learn image-text semantic relation. \newline
Many previous approaches use ranking loss (the training objective is minimizing distance of same image-text pair and maximizing distance of different image-text pair in multi-modal space) for multi-modal embedding. Karpathy's work\cite{karpathy2015deep} use R-CNN\cite{girshick2015fast} for image feature and BRNN\cite{schuster1997bidirectional} for text feature and apply ranking loss. And Some approaches (\cite{frome2013devise},\cite{Wang_2016_CVPR}) use VGG-net for image feature extracting and use neural-language-model for text feature extracting and apply ranking loss or triplet ranking loss. \newline
Some other approaches use deep generative model\cite{sohn2014improved} or DBM (Deep Boltzmann Machine)\cite{srivastava2012multimodal} for multi-modal representation learning. In these methods, they intentionally miss one modality feature and generate missed feature from other modality feature to learn relation of different modalities. Therefore they also use image-text pair information and the process is complicate and not intuitive.

\subsection{Adversarial Network}
Adversarial network concept has started from GAN (Generative Adversarial Network) \cite{goodfellow2014generative}. This concept showed great results for several different tasks. For example, DCGAN (Deep Convolutional Generative Adversarial Network) \cite{DBLP:journals/corr/RadfordMC15} drastically improve generated image quality. And text-conditioned DCGAN \cite{DBLP:conf/icml/ReedAYLSL16} generate related image from text. 
Besides image generation, some approach \cite{icml2015_ganin15} apply adversarial learning concept to domain adaptation field with gradient reversal layer. They did domain adaptation from pre-trained image classification network to semantically similar but visually different domain (e.g. edge image, low-resolution image) image target. For this, they set category predictor and domain classifier, which do adversarial learning, so network's feature trained for category discriminative and domain invariant property.

\subsection{Domain Adaptation and Covariate Shift}
Covariate shift is a primary assumption for domain adaptation field, which assumes that source domain and target domain have same labelling function (same semantic feature or information) but mathematically different distribution form. There was theoretical work about domain adaptation within covariate shift relation source and target domain\cite{adel2015probabilistic}. And we assume that image and text are also in covariate shift relation. We assume image and text have same semantic information (labeling function) but have different distribution form. So our multi-modal embedding process is adapting those distributions as same and retain semantic information at the same time. 

\section{Multi-Modal Representation Learning by Adversarial Backpropagation}
\begin{figure*}[ht]
    \includegraphics[width=\textwidth]{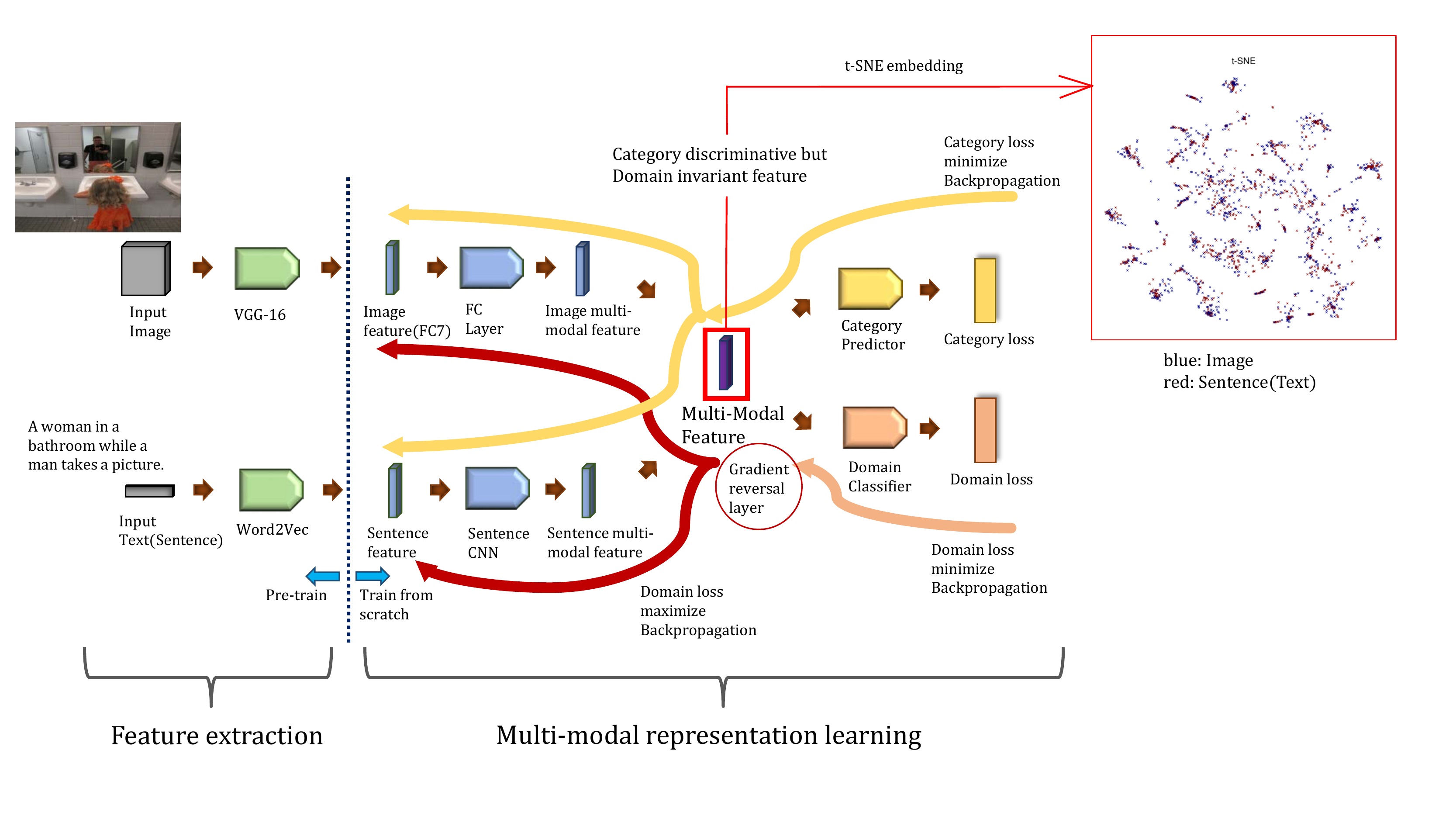}
    \caption{Our adversarial multi-modal embedding network structure. Arrows mean backpropagation of each loss. Feature extraction part is pre-trained by large data set and multi-modal representation learning part is trained with MS COCO dataset's from scratch. Right part of figure is t-SNE embedding of multi-modal feature. (best viewed with color)}
    \label{fig:model}
\end{figure*}
Now, we detail our proposed novel method and model (Figure \ref{fig:model}), multi-modal representation learning by adversarial backpropagation. We use MS COCO dataset \cite{lin2014microsoft}.

\subsection{Network Structure}
Our network structure (Figure \ref{fig:model}) is divided into two parts: feature extraction and multi-modal representation learning. The former part aims at transforming each modality signal into feature. The latter part is devised to embed each feature representation into single (multi-modal) space.\newline
For the representation of visual features, we use VGG16 \cite{DBLP:journals/corr/SimonyanZ14a} which is pre-trained on ImageNet\cite{deng2009imagenet}. To extract image features, we re-size an image to the size $255\times 255$ and crop $224 \times 224$ patches from the four corners and the center. Then the 5 cropped area are flipped to get total 10 patches. We extract fully-connected features (FC7) from each patch, and average them to get a single feature. \newline
To represent sentences, we use Word2Vec \cite{mikolov2013distributed}, which embeds word into 300-dimensional semantic space. In feature extraction process, words in a sentence are converted into Word2Vec vectors, each of which is a 300-dimensional vector. If a sentence contains $N$ words, we get a feature whose size is $N \times 300$. We add zero padding to the bottom row of the feature to fix its size. Since the maximum length of a sentence in MS COCO dataset \cite{lin2014microsoft} is 59, we set the feature size to $59 \times 300$. \newline
After extracting features from each modality, the multi-modal representation learning process follows.  \newline
For an image feature $X \in {\rm I\!R}^{4096}$ and a sentence feature $Y \in {\rm I\!R}^{59 \times 300}$, we apply two transformations $f^{img}$ and $f^{sent}$ for images and sentences respectively, to embed two features into a single $D$-dimensional space. That is, $f^{img}(X) \in {\rm I\!R}^{D}$ and $f^{sent}(Y) \in {\rm I\!R}^{D}$ are satisfied.\newline
For embedding image feature $X \in {\rm I\!R}^{4096}$, we use two fully connected layers with ReLU\cite{nair2010rectified} activation. Since sentence feature $Y \in {\rm I\!R}^{59 \times 300}$ is 2-dimensional, we apply textCNN \cite{DBLP:conf/emnlp/Kim14} to $Y$, to make it possible for $Y$ to be embedded in the space. At the end of each feature embedding network, we use batch-normalization\cite{DBLP:journals/corr/IoffeS15} and L2-normalization respectively. And we apply dropout\cite{srivastava2014dropout} for all fully-connected layers.\newline
The embedding process is regulated by two components -- category predictor and domain classifier with gradient reversal layer, which is a similar concept to that of \cite{icml2015_ganin15}. The category predictor regulates the features on the multi-modal space, in such a way that multi-modal features are discriminative enough to be classified into the valid categories. Meanwhile, the domain classifier with gradient reversal layer makes the multi-modal features being invariant to their domain.\newline

\subsection{Domain Classifer with Gradient Reversal Layer}
We adopt the concept from \cite{icml2015_ganin15}.\newline
GRL (Gradient Reversal Layer) is a layer in which backward pass is reversing gradient values. For a layer's input $X$, the output $Y$, and the identity matrix $I$, forward pass is shown on equation \ref{eq:grl_forward}. In backward pass, for the loss of the network $L$, the gradient subject to $X$ is shown on equation \ref{eq:grl_backward}. $\lambda$ is adaptation factor which is the amount of domain invariance we want to achieve at a point of training.
\begin{equation}
\label{eq:grl_forward}
Y = IX
\end{equation}
\begin{equation}
\label{eq:grl_backward}
\frac{\partial L}{\partial X} = - \lambda\frac{\partial L}{\partial Y}
\end{equation}

The domain classifier is a simple neural network that has two fully-connected layers, with the last sigmoid layer that determines the domain of features in the multi-modal embedding space. That is, it is trained in a way that it discriminates the difference between features from two domains.\newline
However, since the GRL reverses the gradient, feature embedding networks are trained to generate features whose domains are difficult to be determined by the domain classifier. This makes adversarial relationship between the embedding network and the domain classifier. Consequently, domain-invariant features can be generated by the multi-modal embedding networks.\newline

\subsection{Optimization using Backpropagation}
For the calculation of network loss, we sum the losses of two ends -- category predictor and domain classifier. We use sigmoid cross entropy loss for the two ends.
For calculating joint gradient of the category predictor and the domain classifier, the two gradients are added, which is shown in the equation below.
\begin{equation}
\label{eq:f_gradient}
\frac{\partial L}{\partial Y} = \frac{\partial L_c}{\partial Y}  -\lambda\frac{\partial L_d}{\partial Y}
\end{equation}
where $L_c$ and $L_d$ is the error of category predictor and domain classifier respectively, $Y$ is the output of the last feature embedding layer, and $\lambda$ is the adaptation factor.\newline
We use Adam optimizer \cite{DBLP:journals/corr/KingmaB14} for training with relatively small learning rate $10^{-5}$. That's because of the empirical difficulty of generating domain-invariant features with regular learning rate.\newline

\subsection{Adversarial Learning Procedure}
To achieve domain invariant feature, we use domain classifier and gradient reversal layer \cite{icml2015_ganin15}. And we should properly schedule $\lambda$  (adaptation factor) value from 0 to some positive value. Because, at the first stage of training, domain classifier should become smart in advance for adversarial learning process. And with $\lambda$ value increasing, domain classifying with multi-modal feature become difficult and domain classifier become smarter to classify it correctly. In our experiment, it turns out that proper $\lambda$ scheduling is important to achieve domain invariant feature. After exploring many scheduling methods, we find below schedule scheme is optimal, which is exactly the same scheduling as \cite{icml2015_ganin15}. In the equation, $p$ is the fraction of current step in max training steps.
\begin{equation}
\label{eq:lambda_schedule}
\lambda = \frac{2}{1 + exp(-10*p)} - 1
\end{equation}

\subsection{Normalization}
We used batch normalization\cite{DBLP:journals/corr/IoffeS15} and L2-normalization for normalizing image and text feature distribution just before the multi-modal feature layer. In our experiment, without proper normalization, it seems to be trained well (loss value decreases gently and classification accuracy is fine) but when checking the t-SNE\cite{van2014accelerating} embedding and search result, we can recognize that image and text feature distribution is collapsed just for achieving domain invariant feature (collapsed means distance between features going to zero). So proper normalization process is important to achieve domain invariant and also well distributed multi-modal feature.

\section{Result}
\subsection{t-SNE}
\begin{figure}[H]
    \includegraphics[width=\linewidth]{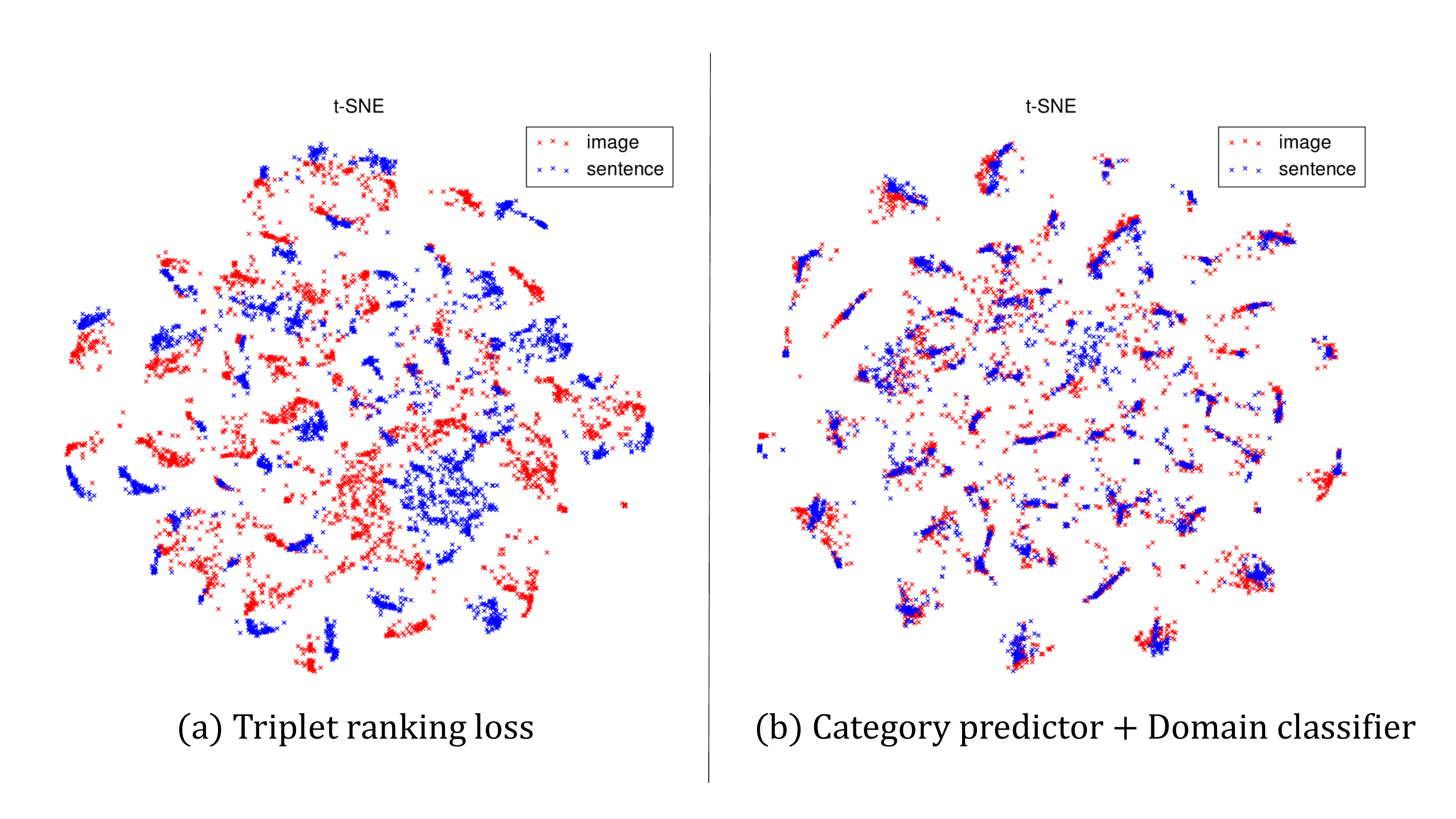}
    \caption{t-SNE embedding with multi-modal feature from image and text. (a) and (b) trained with MS COCO train+validation set . Embedded 5000 data samples come from MS COCO test set. (a): Triplet Ranking Loss (b): Category Predictor + Domain Classifier (our model) (red: image, blue: text) (best viewed with color)}
    \label{fig:tsne}
\end{figure}

Above Figure \ref{fig:tsne} is a t-SNE\cite{van2014accelerating} embedding result of computed multi-modal features from MS COCO test set's 5000 images and sentences. (a) is result of trained with triplet ranking loss (the training objective is minimizing distance of same image-text pair and maximizing distance of different image-text pair in multi-modal space) which exploits image-text pair relation. For implementing, we consult wang's\cite{Wang_2016_CVPR} work. We use FV-HGLMM\cite{klein2015associating} for sentence representation, pre-trained VGG16 for image representation and two-branch fully-connected layers for multi-modal embedding, which is same as wang's did. Difference is they use complex data sampling scheme and we use random data sampling at training stage. (b) is result of trained with category predictor and domain classifier which uses our model. In (a), Image and text feature distributions are not well mixed, which means image and text multi-modal feature's distributions in multi-modal space are not similar. Image and text multi-modal features are not overlapped in multi-modal space. It means, semantically similar image and text are not embedded to near points of multi-modal space. But in (b), our result, we can get well mixed with domains image-text multi-modal feature distribution. Image and text are overlapped in multi-modal space and also distributed enough for being discriminated. It means, image and text has similar distribution in multi-modal space. We think difference of (a) and (b) comes from difference of training objective. Because our model(b) trained for hard to classify domain (image, text) of the multi-modal feature and triplet ranking loss(a) is trained for adhering same image-text pair and pushing different image-text pair. And result means triplet ranking loss not adapt image and text to same distribution in multi-modal space. So this result shows our model's training objective is more suitable for learning well mixed with domains and also well distributed multi-modal feature than other methods.

\subsection{Classification}
\begin{table}[H]
\begin{tabular}{|c||c|c|c|}
    \hline
    Mode & Precision & Recall & F1-score \\
    \hline\hline
    Image only & 0.78 & 0.81 & 0.79 \\
    \hline
    Text only & 0.78 & 0.75 & 0.76 \\
    \hline
    \textbf{Image+Text(m)} & 0.72 & 0.68 & 0.70 \\
    \hline
\end{tabular}
\caption{category classification result of MS COCO val set with various modes. "Image only" means just use vgg-net and category predictor, "Text only" means just use word2vec, TextCNN and category predictor. So, "Image only" and "Text only" modes don't include domain classifier and gradient reversal layer. "Image+Text(m)" is our multi-modal network model (Figure \ref{fig:model}).}
\label{table:classification}
\end{table}
Category classification precision/recall result (Table\ref{table:classification}) shows that after multi-modal embedding, multi-modal feature's (Image+Text(m)) category discriminative power decreases a little, compared to before multi-modal embedding (Image only, Text only), even though it is domain invariant feature. It means our model can adapt image and text feature distribution to multi-modal distribution without large information loss. 

\subsection{Search}
\begin{figure}[H]
    \includegraphics[width=\linewidth]{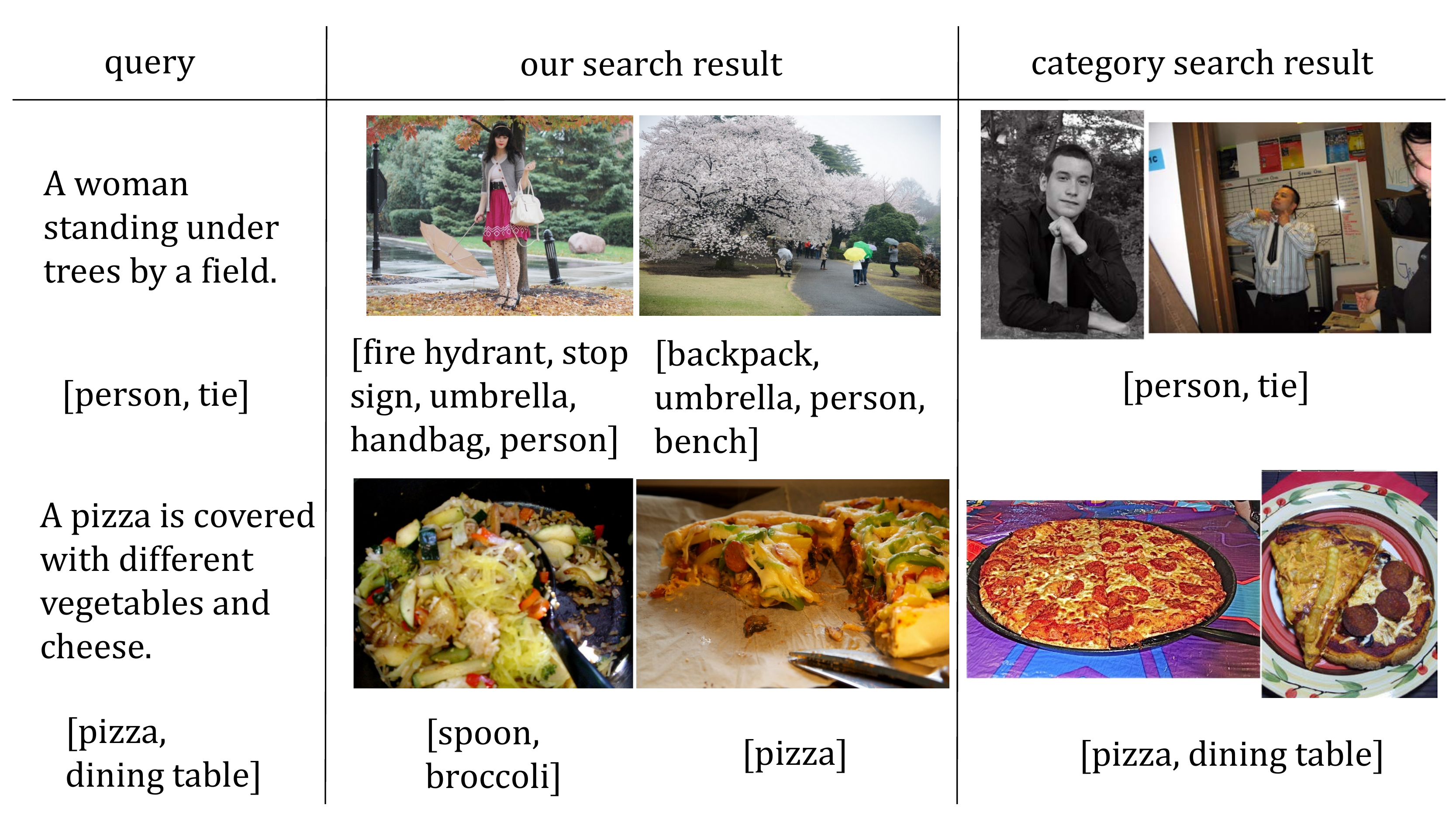}
    \caption{comparison of our sentence to image search result and category based sentence to image search result. We can see our search model not only just care about category information and more universal semantic information even though it just trained for category discriminative multi-modal feature.}
    \label{fig:search_compare}
\end{figure}
\begin{figure*}[!ht]
    \includegraphics[width=\textwidth]{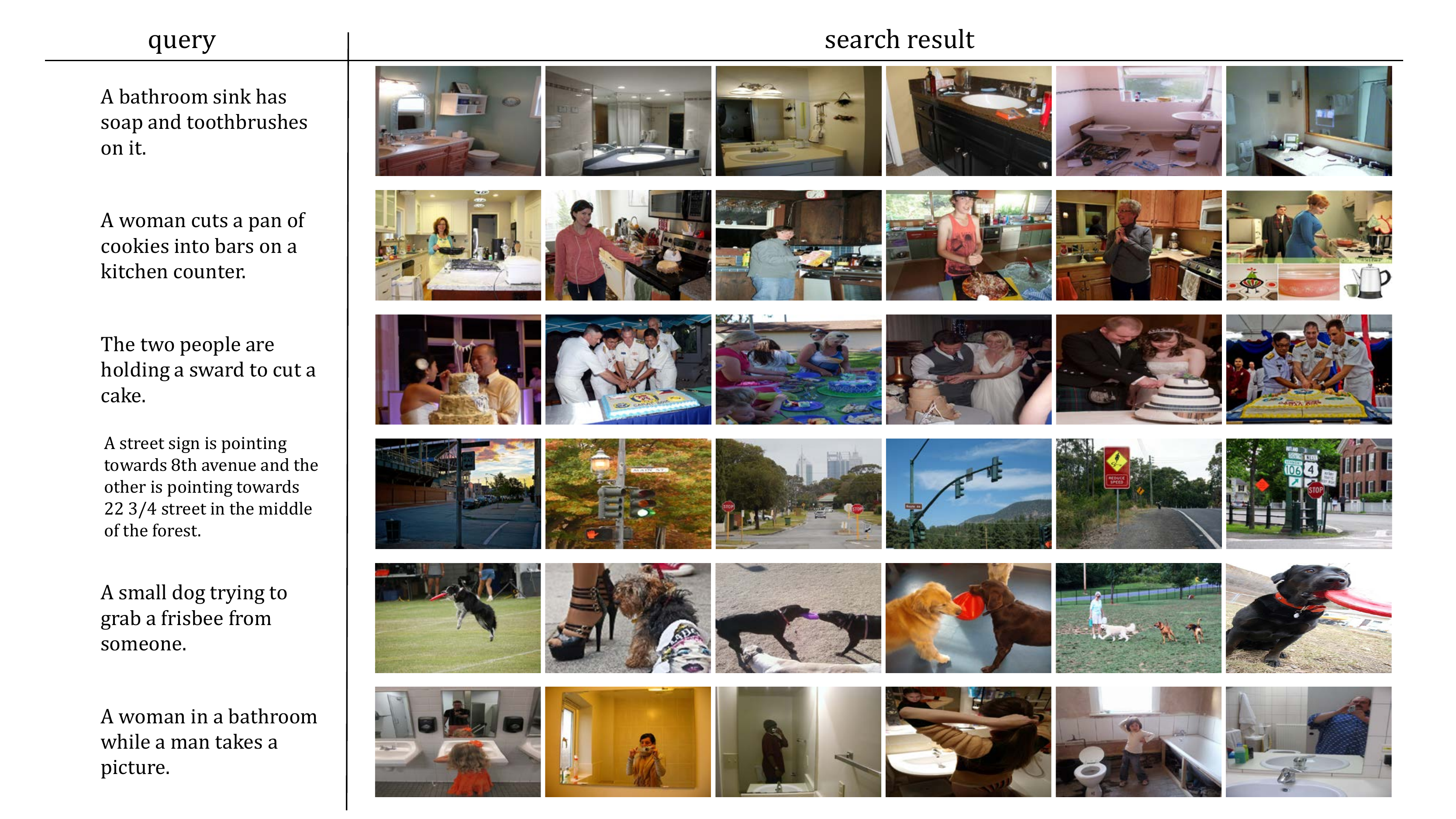}
    \caption{sentence to image search result (top 6) example from 40504 number of MS COCO validation set.}
    \label{fig:search_result}
\end{figure*}
\begin{table*}[!ht]
\centering
\begin{tabular}{llllllll}
\hline
\multicolumn{1}{|r|}{}                                                       & \multicolumn{1}{l|}{Model}                          & \multicolumn{3}{l|}{Image-to-Sentence}                                            & \multicolumn{3}{l|}{Sentence-to-Image}                                            \\ \hline
\multicolumn{1}{|l|}{}                                                       & \multicolumn{1}{l|}{}                               & \multicolumn{1}{l|}{R@1}  & \multicolumn{1}{l|}{R@5}  & \multicolumn{1}{l|}{R@10} & \multicolumn{1}{l|}{R@1}  & \multicolumn{1}{l|}{R@5}  & \multicolumn{1}{l|}{R@10} \\ \hline
\multicolumn{1}{|l|}{\multirow{4}{*}{\shortstack{Using image-text \\ pair information}}}     & \multicolumn{1}{l|}{DVSA\cite{karpathy2015deep}}                           & \multicolumn{1}{l|}{38.4} & \multicolumn{1}{l|}{69.9} & \multicolumn{1}{l|}{80.5} & \multicolumn{1}{l|}{27.4} & \multicolumn{1}{l|}{60.2} & \multicolumn{1}{l|}{74.8} \\ \cline{2-8} 
\multicolumn{1}{|l|}{}                                                       & \multicolumn{1}{l|}{m-RNN-vgg\cite{mao2014deep}}                      & \multicolumn{1}{l|}{41.0} & \multicolumn{1}{l|}{73.0} & \multicolumn{1}{l|}{83.5} & \multicolumn{1}{l|}{29.0} & \multicolumn{1}{l|}{42.2} & \multicolumn{1}{l|}{77.0} \\ \cline{2-8} 
\multicolumn{1}{|l|}{}                                                       & \multicolumn{1}{l|}{mCNN (ensemble)\cite{ma2015multimodal}}                 & \multicolumn{1}{l|}{42.8} & \multicolumn{1}{l|}{73.1} & \multicolumn{1}{l|}{84.1} & \multicolumn{1}{l|}{32.6} & \multicolumn{1}{l|}{68.6} & \multicolumn{1}{l|}{82.8} \\ \cline{2-8} 
\multicolumn{1}{|l|}{}                                                       & \multicolumn{1}{l|}{structure preserving\cite{Wang_2016_CVPR}} & \multicolumn{1}{l|}{50.1} & \multicolumn{1}{l|}{79.7} & \multicolumn{1}{l|}{89.2} & \multicolumn{1}{l|}{39.6} & \multicolumn{1}{l|}{75.2} & \multicolumn{1}{l|}{86.9} \\ \hline
\multicolumn{1}{|l|}{\multirow{2}{*}{\shortstack{  Using category information }}} & \multicolumn{1}{l|}{\textbf{our model (TextCNN)}}           & \multicolumn{1}{l|}{13.6}   & \multicolumn{1}{l|}{39.6}   & \multicolumn{1}{l|}{54.6}   & \multicolumn{1}{l|}{10.3}   & \multicolumn{1}{l|}{35.5}   & \multicolumn{1}{l|}{55.5}   \\ \cline{2-8} 
\multicolumn{1}{|l|}{}                                                       & \multicolumn{1}{l|}{\textbf{our model (FV-HGLMM)}}           & \multicolumn{1}{l|}{14.3}   & \multicolumn{1}{l|}{40.5}   & \multicolumn{1}{l|}{55.8}   & \multicolumn{1}{l|}{12.7}   & \multicolumn{1}{l|}{39.0}   & \multicolumn{1}{l|}{57.2}   \\ \hline
                                                                             &                                                     &                           &                           &                           &                           &                           &                          
\end{tabular}
\caption{Bidirectional retrieval recall@K results on MS COCO 1000-image test set. our model (TextCNN) and our model (FV-HGLMM) use TextCNN and FV-HGLMM for text representation respectively.}
\label{table:recallk}
\end{table*}
We build our sentence to image search system with 40504 number of MS COCO validation set which is never seen at training stage. We train with 82783 images (train set) and test for 40504 images (val set). And simply do k-nearest-neighbor search in multi-modal space with computed multi-modal feature. \newline
Figure \ref{fig:search_compare} shows comparison of our search result and category based search result. You can see the sentence query has more semantic information than its category label. And category based search cannot exploit that semantic information but our search system can exploit that semantic information of sentence query. In figure \ref{fig:search_compare}, we can see our search system finds several objects which is not contained in category information but exists in sentence query. In 1st row of figure \ref{fig:search_compare}, our search system rightly catches information ``woman standing under trees by a field" from sentence even though it was just trained to predict [person, tie] from sentence and image at training time. It means our multi-modal embedding process didn't remove universal information extracted from word2vec and VGG16. And also match image and text semantically relevant feature during multi-modal embedding process. \newline
In 2nd row of figure \ref{fig:search_compare}, our search system thinks that most similar image with the query is food image which that category is [spoon, broccoli], which is not overlapped with query's category. But interestingly, in human's semantic perspective, we can recognize they have similar semantic information. (``covered with different vegetables and cheese.'') \newline
In Figure \ref{fig:search_result} (next page), you can see more various search results from our multi-modal search system.

For benchmark of search system, we did recall@K evaluation (table \ref{table:recallk}, next page) with sentence-to-image and image-to-sentence retrieval. For this, we used karpathy's data split scheme\cite{karpathy2015deep}. Compare to state-of-the-art results, our model's performance is relatively low. We think, the major reason is, previous models trained for adhering image-text pair and pushing different image-text pair in multi-modal space and recall@K evaluate query's pair appeared or not in retrieval result. So, even if search result is semantically reasonable (figure \ref{fig:search_result}), if query's pair not appear in retrieval result, recall@K can be low. So we think, this metric is not fully appropriate to assess search quality. But for comparison, we also did recall@K experiment.
\section{Conclusion}
We have proposed a novel approach for multi-modal representation learning which uses adversarial backpropagation concept. Our method does not require image-text pair information for multi-modal embedding but only uses category label. In contrast, until now almost all other methods exploit image-text pair information to learn semantic relation between image and text feature.\newline
Our work can be easily extended to other multi-modal representation learning (e.g. sound-image, sound-text, video-text). So our method's future work will be extending this method to other multi-modal case.

\bibliographystyle{icml2013}

\end{document}